\documentclass[letterpaper, 10 pt, conference]{ieeeconf}  

\IEEEoverridecommandlockouts                              

\usepackage{graphicx} 
\usepackage{amsmath} 
\usepackage{amssymb}  
\usepackage{booktabs}
\usepackage{multirow}

\title{\LARGE \bf
MAP-VLA: Memory-Augmented Prompting for Vision-Language-Action Model in Robotic Manipulation
}

\author{Runhao Li$^{1}$, Wenkai Guo$^{1}$, Zhenyu Wu$^{3}$, Changyuan Wang$^{4}$, Haoyuan Deng$^{1}$, \\Zhenyu Weng$^{5}$, Yap-Peng Tan$^{2,1}$, Ziwei Wang$^{1,*}$
\thanks{$^{1}$Nanyang Technological University, Singapore, Singapore}
\thanks{$^{2}$VinUniversity, Hanoi, Vietnam}
\thanks{$^{3}$Beijing University of Posts and Telecommunications, Beijing, China}
\thanks{$^{4}$Tsinghua University, Beijing, China}
\thanks{$^{5}$South China University of Technology, Guangzhou, China}
\thanks{$^{*}$Corresponding author: ziwei.wang@ntu.edu.sg}
}

\begin{document}

\maketitle
\thispagestyle{empty}
\pagestyle{empty}

\begin{abstract}

Pre-trained Vision-Language-Action (VLA) models have achieved remarkable success in improving robustness and generalization for end-to-end robotic manipulation. However, these models struggle with long-horizon tasks due to their lack of memory and reliance solely on immediate sensory inputs.
To address this limitation, we propose Memory-Augmented Prompting for Vision-Language-Action
model (MAP-VLA), a novel framework that empowers pre-trained VLA models with demonstration-derived memory prompts to augment action generation for long-horizon robotic manipulation tasks. To achieve this, MAP-VLA first constructs a memory library from historical demonstrations, where each memory unit captures information about a specific stage of a task.
These memory units are implemented as learnable soft prompts optimized through prompt tuning.
Then, during real-time task execution, MAP-VLA retrieves relevant memory through trajectory similarity matching and dynamically integrates it into the VLA model for augmented action generation. Importantly, this prompt tuning and retrieval augmentation approach operates as a plug-and-play module for a frozen VLA model, offering a lightweight and flexible solution to improve task performance. Experimental results show that MAP-VLA delivers up to 7.0\% absolute performance gains in the simulation benchmark and 25.0\% on real robot evaluations for long-horizon tasks, surpassing the current state-of-the-art methods.

\end{abstract}

\section{INTRODUCTION}

Robotic manipulation remains a long-standing challenge in embodied artificial intelligence, requiring a robot to perceive complex scenes, understand task goals, and generate multi-step control actions. 
Traditional approaches~\cite{kroemer2021review} often rely on task-specific pipelines or require substantial engineering and manual tuning, which limits their adaptability and scalability.
In response to these limitations, Vision-Language-Action (VLA) models~\cite{zitkovich2023rt,kim2024openvla,black2024pi_0,pertsch2025fast,wen2025tinyvla,zhen20243d,zhao2025cot} have recently emerged as a compelling new paradigm for developing generalist robotic manipulation policies. These models build upon the success of large-scale vision-language foundation models by extending their capabilities into the action domain.
VLA models acquire broad knowledge about the world from vision-language pre-training and learn to map raw visual observations and natural language instructions directly to robot actions through end-to-end training on diverse robotic datasets~\cite{o2024open,khazatsky2024droid,ebert2021bridge}.
By unifying visual perception, language understanding, and action generation into a single policy, they offer a powerful foundation for general-purpose robotic manipulation.

\begin{figure*}[h]
\centering
\includegraphics[width=1.98\columnwidth]{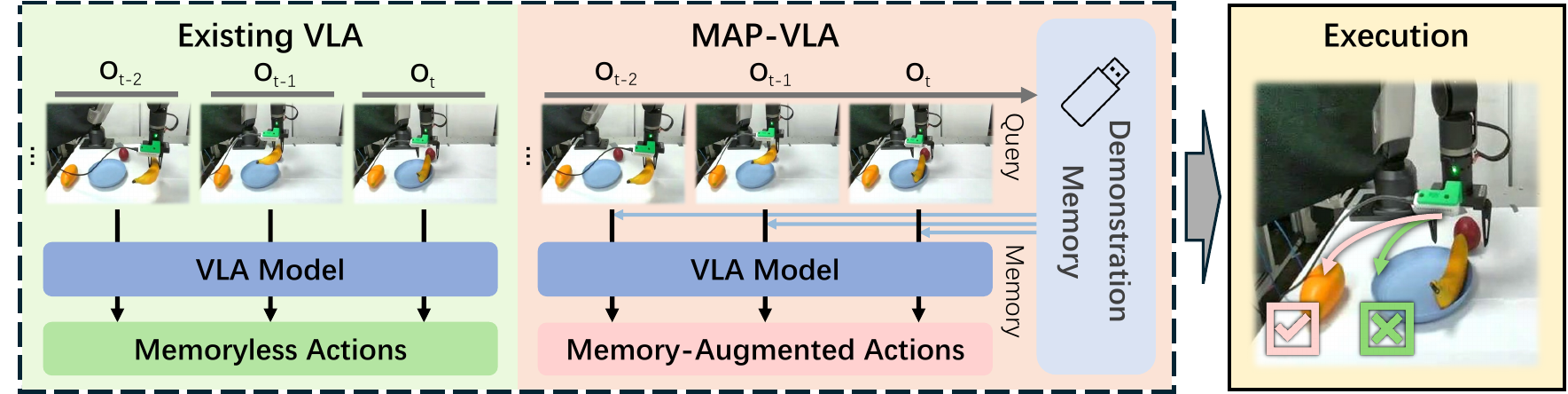} 
\caption{Simplified execution pipeline of existing VLA methods and MAP-VLA.}
\label{fig:intro}
\end{figure*}

Despite the advantages mentioned above, current VLA models have a key limitation: they fail to leverage historical memory at task execution. Once a VLA model is trained, it relies solely on immediate sensory inputs to decide the subsequent actions. The rich experience contained in the training demonstrations is not explicitly accessible during execution as historical memory, and is only used to train the model parameters offline. In other words, the robot lacks episodic memory; it cannot directly recall “how an expert accomplished a similar stage” when it is performing one. This stateless execution is suboptimal, especially for long-horizon tasks, and the robot may deviate from the intended trajectory in challenging situations that an expert has encountered before. Existing approaches have explored in-context learning for robots~\cite{fu2024context,yin2024context,vosylius2024instant,zhu2024incoro}, for instance, by providing a few demonstration episodes as part of the input to a policy model. However, such solutions either require extensive training tailored to specific architectures and input configurations, or concentrate on LLMs without applicability to pre-trained VLA models. In general, existing VLA models do not natively support the use of historical data for action generation at test time, leaving a gap between how humans solve long-horizon tasks, by recalling past memory, and how these models operate. This motivates the need for approaches that can endow VLA models with a lightweight form of episodic memory.

\begin{figure*}[t]
\centering
\includegraphics[width=1.98\columnwidth]{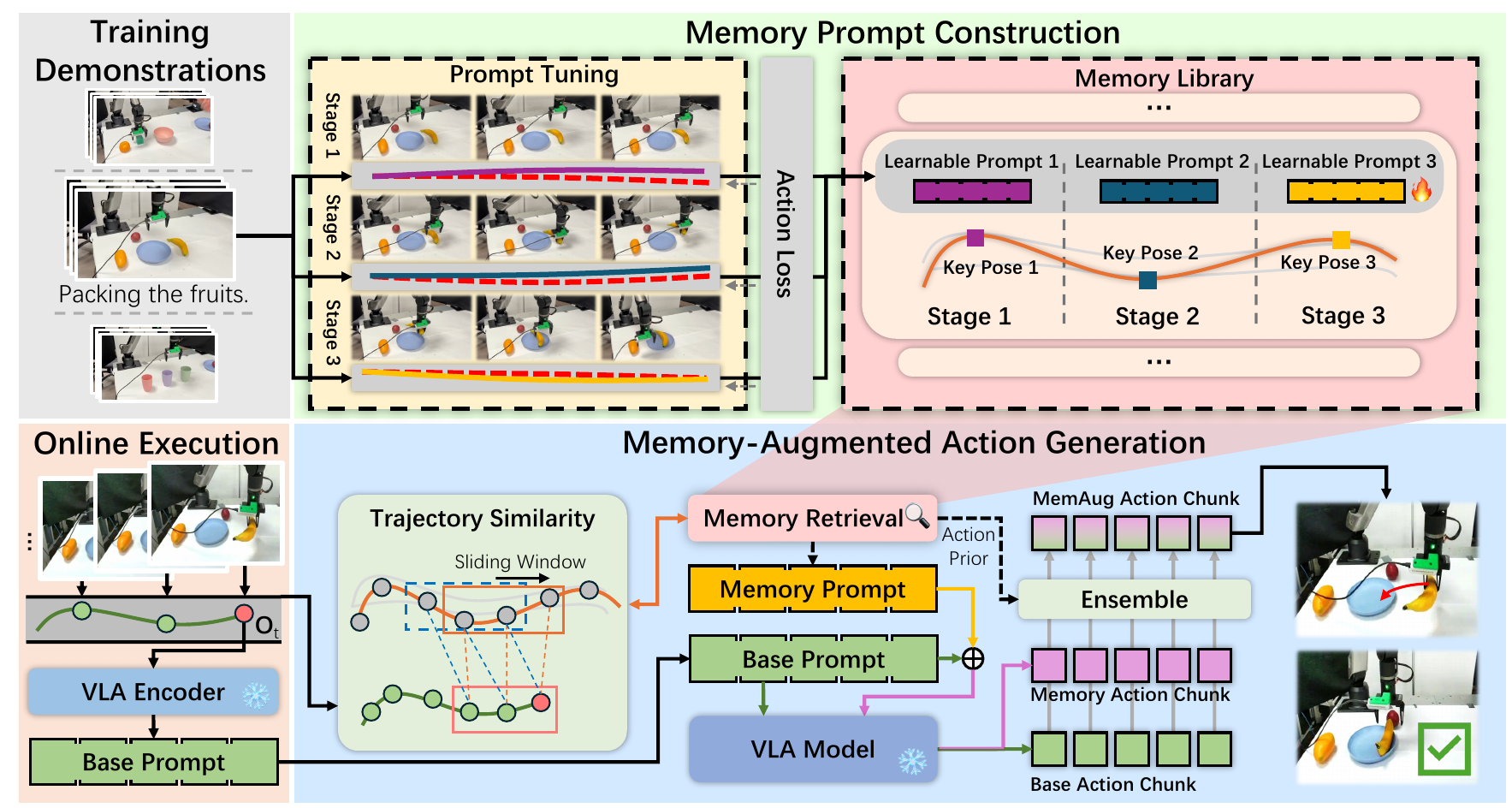} 
\caption{The framework of MAP-VLA. Our method augments a frozen pre-trained VLA model with demonstration-derived memory prompts for enhanced action generation during task execution. The Memory Prompt Construction stage encodes stage-specific knowledge from expert demonstrations into a library of memory prompts. The Memory-Augmented Action Generation stage retrieves the memory prompts and augments action generation with memory-aware prompt ensembling.}
\label{fig:framework}
\end{figure*}

In this paper, we present the Memory-Augmented Prompting for Vision-Language-Action
model (MAP-VLA), bridging the gap in current VLA models by enabling dynamic access to demonstration-derived memory. As illustrated in Fig.~\ref{fig:intro}, our insight is to equip a frozen VLA model with a lightweight memory of training demonstrations, which can be retrieved during task execution to provide stage-by-stage guidance. 
To achieve this, we first introduce Memory Prompt Construction, where we segment training trajectories into distinct stages and encode each stage’s memory into a soft prompt via prompt tuning. This process effectively builds an external memory library that captures stage-specific knowledge for guiding the model’s behavior.
Next, we propose Memory-Augmented Action Generation, which retrieves the most relevant stage-specific memory prompt along with the corresponding demonstration actions by comparing the trajectory similarity. The memory prompt and demonstration actions are subsequently utilized for prompt ensembling, which dynamically balances the advantages of the stage-specific memory prompts and the generalized base prompts.
This whole framework, shown in Fig.~\ref{fig:framework}, remains lightweight and flexible without updating the underlying model parameters. Extensive experiments demonstrate that MAP-VLA surpasses the state-of-the-art methods on long-horizon tasks with up to 7.0\% absolute gains in the simulation benchmark and 25.0\% in real robot evaluations.
The main contributions of this work can be summarized as follows:
\begin{itemize}
\item We propose MAP-VLA, a novel framework that augments a pre-trained VLA model with demonstration-derived memory prompts. This framework operates on prompt tuning and retrieval-augmented generation to improve task adaptation, without requiring any modification to the model’s internal weights. 
\item We introduce Memory Prompt Construction (MPC), which encodes stage-specific memory from expert demonstrations into a library of prompts to provide episodic task memory. We also develop Memory-Augmented Action Generation (MAAG), which enables memory retrieval and dynamic memory-aware prompt ensembling to augment action generation during real-time task execution.
\item Extensive experiments show that MAP-VLA consistently outperforms the state-of-the-art methods for long-horizon manipulation tasks, achieving up to 7.0\% absolute gains in the simulation benchmark and 25.0\% in real robot evaluations.
\end{itemize}

\section{RELATED WORK}
\subsection{Vision-Language-Action Models}
Recent advances in robot learning have introduced VLA models~\cite{zitkovich2023rt,kim2024openvla,black2024pi_0,pertsch2025fast,wen2025tinyvla,zhen20243d,zhao2025cot} pre-trained on large datasets~\cite{o2024open,khazatsky2024droid,ebert2021bridge} that unify visual perception, natural language, and robot control. A prominent example is RT-2 (Robotic Transformer 2)~\cite{zitkovich2023rt}, which introduced the VLA paradigm by integrating an Internet-scale vision-language model into an end-to-end policy. RT-2 represents robot actions as textual tokens and co-trains on both web data and robot demonstrations, yielding a single model that maps images to actions while enjoying the semantic richness of language pre-training. Building on this idea, OpenVLA~\cite{kim2024openvla} is a 7B-parameter open-source VLA model trained on nearly one million real-world robot episodes. It demonstrated efficient fine-tuning for new tasks and embodiments, significantly outperforming prior imitation learning methods (e.g. diffusion policies) on multi-task benchmarks. The $\pi_0$ model~\cite{black2024pi_0} proposes a novel flow-matching VLA architecture built on top of a pre-trained vision-language model. It employs a diffusion-like flow matching strategy for action generation, enabling controlling high-frequency, highly dexterous skills (up to 50Hz control rates) that are challenging for earlier autoregressive Transformers. 
However, despite differences in architecture and training, these methods share a common limitation: they rely solely on immediate sensory inputs and predefined task instructions, without explicitly utilizing the rich historical memory embedded in expert demonstrations during task execution.

\subsection{Prompt Tuning}
As foundation models became prevalent, prompt tuning~\cite{lester2021power,zhou2022learning,zhou2022conditional,khattak2023maple,yao2024tcp,jia2022visual} emerged as a lightweight method to adapt them to downstream tasks without full fine-tuning. In natural language processing, the literature~\cite{lester2021power} introduced prompt tuning as learning a set of continuous “soft prompt” vectors that are prepended to the input, conditioning a frozen language model to perform a new task. Instead of hand-crafting a textual prompt or updating millions of model weights, prompt tuning optimizes a small embedding that steers the model’s behavior via its inputs. These soft prompts are learned through standard backpropagation on the task objective, effectively encoding the task information in the input space. Remarkably, this approach can achieve performance on par with fine-tuning the entire model, especially as model size grows. Prompt-based adaptation has also proven effective in vision-language models~\cite{zhou2022learning,zhou2022conditional,khattak2023maple,yao2024tcp}. For instance, CoOp~\cite{zhou2022learning} applies soft prompt tuning to CLIP~\cite{radford2021learning}, a vision-language model, for image recognition tasks. CoOp replaces manual prompt engineering (e.g., finding the right wording like “a photo of a [class]”) with learnable prompt embeddings, while freezing the entire CLIP model. Even with only a few training examples, the learned prompts can significantly improve accuracy over handcrafted prompts, and they transfer well to new domains. More broadly, prompt tuning is well-suited for VLA models due to their structural similarity with large language and vision-language models, making it a natural adaptation strategy.

\section{METHODOLOGY}

\subsection{Preliminaries and Overview}

\textbf{Vision-Language-Action pre-training.}  
VLA models are policies pre-trained on large-scale demonstrations to map multi-modal observations to robot actions. 
Each demonstration consists of a sequence of observation-action pairs $\{ \mathbf{o}_t, \mathbf{a}_t \}_{t=1}^{n}$, where each observation $\mathbf{o}_t = [\mathbf{I}_t^1, \mathbf{I}_t^2, \ell_t, \mathbf{s}_t]$ includes an overview image $\mathbf{I}_t^1$, a wrist image $\mathbf{I}_t^2$, a language token sequence $\ell_t$, and the robot state $\mathbf{s}_t$. The corresponding action at time $t$ is represented by $\mathbf{a}_t$.
Our method is based on the pre-trained $\pi_0$~\cite{black2024pi_0} which uses $\mathbf{o}_t$ to predict an action chunk of $H$ future actions $\mathbf{A}_t = [\mathbf{a}_t, \dots, \mathbf{a}_{t+H-1}]$ via a conditional flow matching process. The learning objective can be summarized as follows:
\begin{equation}
L^\tau(\theta) = \mathbb{E}_{p(\mathbf{A}_t | \mathbf{o}_t),\, q(\mathbf{A}^\tau_t | \mathbf{A}_t)} \left\| \mathbf{f}_\theta(\mathbf{A}^\tau_t, \mathbf{o}_t) - \mathbf{u}(\mathbf{A}^\tau_t | \mathbf{A}_t) \right\|^2,
\label{eq:pretrain}
\end{equation}
where $\tau \in [0,1]$ is the flow matching timestep, and $\mathbf{A}_t^\tau = \tau \mathbf{A}_t + (1 - \tau)\epsilon$ is a noisy interpolation with $\epsilon \sim \mathcal{N}(0, \mathbf{I})$. The model $\mathbf{f}_\theta(\mathbf{A}^\tau_t, \mathbf{o}_t)$ predicts the denoising vector field $\mathbf{u}(\mathbf{A}^\tau_t | \mathbf{A}_t) = \epsilon - \mathbf{A}_t$.
While $\pi_0$ and other VLA models excel at learning generalizable mappings from diverse demonstrations, they often struggle in long-horizon tasks due to the absence of memory mechanisms. 

\textbf{Methodology overview}. To overcome this, we introduce a memory-augmented framework that enhances VLA models for better long-horizon task performance. Our framework comprises two components: (i) Memory Prompt Construction (MPC), which encodes stage-level memory into soft prompts derived from expert demonstrations, and (ii) Memory-Augmented Action Generation (MAAG), which retrieves and dynamically integrates these prompts to augment real-time action generation. We detail each component in Section~\ref{sec:pmc} and Section~\ref{sec:maag}, respectively.


\subsection{Memory Prompt Construction}
\label{sec:pmc}

\textbf{Stage segmentation and alignment}.
We begin by partitioning each demonstration into stage segments that enable targeted memory supervision for precise action generation. To identify meaningful task stage boundaries, we first select a well-performed demonstration as reference and extract its key poses that mark salient transitions such as grasps or directional changes from the end-effector's states $\mathbf{S}^{\text{ref}} = [\mathbf{s}^{\text{ref}}_0, \dots, \mathbf{s}^{\text{ref}}_n]$ using the Ramer--Douglas--Peucker (RDP) algorithm~\cite{ramer1972iterative}. The key poses are selected iteratively to preserve essential shape of the trajectory and the stage segmentation $\{ \mathcal{S}_1, \dots, \mathcal{S}_K \}$ is defined by placing each segment boundary at the midpoint between two consecutive key poses, enabling memory to guide the execution throughout the entire key pose.
Furthermore, to ensure stage consistency across demonstrations of the same task, we employ the Dynamic Time Warping (DTW) algorithm~\cite{muller2007dynamic}, a technique that non-linearly aligns two sequences by minimizing their cumulative distance. Specifically, we align each trajectory $\mathbf{S}^{i}$ from the training demonstrations to the RDP-segmented reference trajectory $\mathbf{S}^{\text{ref}}$ by computing the optimal warping path that best matches their temporal progression. This process adjusts for variations in execution speed and duration while preserving the semantic structure of the task. As a result, we obtain $K$ stage-aligned segments ${ \mathcal{S}_1, \dots, \mathcal{S}_K }$ across all demonstrations, ensuring that the $k$-th segment in each demonstration consistently corresponds to the same task stage.

\textbf{Stage-specific prompt tuning}.
Given the segmented and aligned demonstrations, we assign a soft prompt vector for each task stage that encodes stage-specific memory to guide the model in executing the corresponding key pose. At each timestep $t$, the observation $\mathbf{o}_t$ is processed by the image and language encoders to generate a base prompt with $m$ base token embeddings $\mathcal{P}_{\text{base}} = [\mathbf{p}_1, \dots, \mathbf{p}_m]$ from the immediate visual and textual inputs. To incorporate stage-specific memory into large VLA models, we augment the base prompt with a set of learnable soft tokens for each stage. These tokens are optimized via prompt tuning to encode abstract memory derived from training demonstrations, allowing the model to condition its behavior on the current task stage. Specifically, for a given stage $\mathcal{S}_k$, we define a learnable vector sequence $\mathcal{V}_k = [\mathbf{v}^k_1, \dots, \mathbf{v}^k_m]$ as soft prompt. The final stage-specific memory prompt $\mathcal{P}_{k}$ input to the VLA model is computed via element-wise addition:  
\begin{equation}
\left[ \mathcal{P}_k \right]_j = \left[ \mathcal{P}_{\text{base}} \right]_j + \left[ \mathcal{V}_k \right]_j, \quad \forall j = 1,\dots,m.
\label{eq:prompt}
\end{equation}
Since $\mathcal{P}_{\text{base}}$ is conditioned on $\mathbf{o}_t$, which varies within a stage, this additive formulation enables efficient integration of contextual memory without compromising the model’s base capability and real-time adaptability. To encode the stage-specific memory, we optimize $\mathcal{V}_k$ by aligning the model’s predicted action tokens with expert actions using the flow matching loss:
\begin{multline}
\mathcal{V}_k^* = \arg\min_{\mathcal{V}_k} \,
\mathbb{E}_{\substack{p(\mathbf{A}_t \mid \mathbf{o}_t),\, q(\mathbf{A}_t^\tau \mid \mathbf{A}_t)}} \\
\left\| \mathbf{f}_{\theta}(\mathbf{A}_t^\tau, \mathbf{o}_t, \mathcal{V}_k)
 - \mathbf{u}(\mathbf{A}_t^\tau \mid \mathbf{A}_t) \right\|^2,
\quad t \in \mathcal{S}_k.
\label{eq:optimize}
\end{multline}
This process effectively encodes the demonstration memory of a stage into the prompt embeddings to augment action generation.
Finally, by tuning the vector sequences for all stages of a task, we construct the task memory prompts as 
$\left\{ \mathcal{V}_k \right\}_{k=1}^{K}$.
Repeating this process for all tasks yields a memory library, a repository of stage-specific prompts that encapsulates contextual cues as retrievable memory. We also add in the demonstration trajectories $\left\{ \mathbf{S}^{i} \right\}_{i=1}^{N}$ and action sequences $\left\{ \mathbf{A}^{i} \right\}_{i=1}^{N}$ in this library, where $N$ is the total number of training demonstrations. This library is subsequently used to augment action generation in the following component.



\subsection{Memory-Augmented Action Generation}
\label{sec:maag}

\textbf{Memory retrieval}. To effectively retrieve relevant memory during task execution, the system must relate the robot’s ongoing behavior with historical demonstrations. We address this by comparing the similarity between the robot’s ongoing trajectory and the training demonstrations.
Specifically, we define a fixed window size $W$ and denote the last $W$ steps of the robot’s ongoing states up to time~$t$ as $\mathbf{S}^{\text{cur}} = [\mathbf{s}^{\text{cur}}_{t-W+1}, \dots, \mathbf{s}^{\text{cur}}_{t}]$. For each demonstration trajectory $\mathbf{S}^{i}$, we perform a sliding window search to compute the similarity between the recent segment of $\mathbf{S}^{\text{cur}}$ and candidate segments from $\mathbf{S}^{i}$.
For each candidate index $j$ in $\mathbf{S}^{i}$, we compute the $\ell_2$ distance as cost $\mathcal{C}$ between the current trajectory window and the corresponding reference window:
\begin{equation}
\mathcal{C}_{i,j} = \left\| \mathbf{S}^{\text{cur}} - \mathbf{S}^{i}[j - W + 1 : j] \right\|_2.
\label{eq:similarity}
\end{equation}
However, since long-horizon tasks often yield lengthy trajectories, the sliding window search can become computationally expensive, and similar trajectory windows may appear across different task stages. To address this, we adopt a hierarchical strategy by restricting candidate indices $j$ to those whose associated stage label $k^i_{j}$ and the most recently identified stage label $k^{\text{cur}}$ satisfy $|k^i_{j} - k^{\text{cur}}| \leq 1$, thereby ensuring retrieval is limited to neighboring task stages.
Then, we select the reference index $j^*$ and trajectory $i^*$ minimizing the distance:
\begin{equation}
i^*, j^* = \arg\min_{i,j} \; \mathcal{C}_{i,j}.
\label{eq:retrieval}
\end{equation}
The task stage associated with $(i^*, j^*)$ is then used to update the current task stage $k^{\text{cur}}$ and retrieve the corresponding $\mathcal{V}_{k^{\text{cur}}}$ from the memory prompt library. This retrieved prompt reflects the memory of expert behavior for the most similar historical situation encountered from demonstrations. 

\textbf{Memory-aware prompt ensembling}. 
While retrieving memory prompts provides valuable memory from past demonstrations tailored to the current stage of execution, the base prompt, trained over the entire task, provides broader task-level generalization. However, each alone is insufficient. The retrieved memory is vulnerable to retrieval inaccuracies, ambiguous execution stages, and stage misalignment of training demonstrations. Conversely, the base prompt, derived solely from current observations and task instructions, is not affected by such errors but lacks historical grounding and long-horizon memory.
To reconcile these complementary strengths, we introduce a memory-aware prompt ensembling mechanism that dynamically combines the stage specificity of memory prompts with the task generalization capability of base prompts.
At each timestep $t$, the frozen VLA model produces two action predictions: $\mathbf{A}^{\text{base}}_{t}$ using the base prompt $\mathcal{P}_{\text{base}}$, and $\mathbf{A}^{\text{mem}}_{t}$ using the retrieved stage-specific memory prompt $\mathcal{P}_{k^*}$. In parallel, we retrieve reference actions $\mathbf{A}^i_{j} = [\mathbf{a}^i_{j}, \dots, \mathbf{a}^i_{j+H-1}]$ from demonstration $i = i^*$, starting at index $j = j^*$. This action sequence encapsulates valuable action priors about the best-matching demonstration's future actions under a similar situation. However, due to differences between the ongoing task and the historical demonstration, directly executing $\mathbf{A}^i_{j}$ is not feasible. Instead, we use it as an action prior to guide the dynamic weighting between $\mathbf{A}^{\text{base}}_{t}$ and $\mathbf{A}^{\text{mem}}_{t}$. Specifically, we compare these predictions to the retrieved demonstration action $\mathbf{A}^i_{j}$ and quantify a dynamic weighting coefficient $\alpha_t$ computed via softmax normalization:
\begin{equation}
    \alpha_t = \frac{\exp(- \left\| \mathbf{A}^{\text{mem}}_{t} - \mathbf{A}^i_{j} \right\|^2)}{\exp(- \left\| \mathbf{A}^{\text{mem}}_{t} - \mathbf{A}^i_{j} \right\|^2) + \exp(- \left\| \mathbf{A}^{\text{base}}_{t} - \mathbf{A}^i_{j} \right\|^2)}.
\label{eq:weight}
\end{equation}

\begin{table*}[t]
\centering
\small
\caption{Performance comparison on LIBERO-Long simulation benchmark. For each method, the first row reports success rates, and the second row reports standard deviations.}
\label{tab:libero-long}
\begin{tabular}{l|cccccccccc|c}
\toprule
\textbf{Method} & Task1 & Task2 & Task3 & Task4 & Task5 & Task6 & Task7 & Task8 & Task9 & Task10 & Avg \\
\midrule
\multirow{2}{*}{OpenVLA}
& 75.3\% & 30.7\% & 64.0\% & 62.0\% & 68.0\%
& 41.3\% & 52.0\% & 50.0\% & 42.7\% & 54.0\%
& 54.0\% \\
& $\pm$\,8.1\% & $\pm$\,4.2\% & $\pm$\,3.5\% & $\pm$\,2.0\% & $\pm$\,3.5\%
& $\pm$\,3.1\% & $\pm$\,1.2\% & $\pm$\,0.0\% & $\pm$\,6.1\% & $\pm$\,5.3\%
& $\pm$\,0.4\% \\
\midrule
\multirow{2}{*}{$\boldsymbol{\pi_0}$}
& 80.0\% & 38.0\% & 88.0\% & 88.0\% & 90.7\%
& 85.3\% & 81.3\% & 68.7\% & 66.7\% & 77.3\%
& 76.4\% \\
& $\pm$\,4.0\% & $\pm$\,3.5\% & $\pm$\,4.0\% & $\pm$\,4.0\% & $\pm$\,1.2\%
& $\pm$\,2.3\% & $\pm$\,3.1\% & $\pm$\,12.2\% & $\pm$\,11.4\% & $\pm$\,4.6\%
& $\pm$\,2.3\% \\
\midrule
\multirow{2}{*}{MAP-VLA}
& \textbf{92.0\%} & \textbf{42.7\%} & \textbf{96.0\%} & \textbf{90.7\%} & \textbf{93.3\%}
& \textbf{93.3\%} & \textbf{90.7\%} & \textbf{75.3\%} & \textbf{69.3\%} & \textbf{90.7\%}
& \textbf{83.4\%} \\
& $\pm$\,2.0\% & $\pm$\,2.3\% & $\pm$\,2.0\% & $\pm$\,1.2\% & $\pm$\,1.2\%
& $\pm$\,3.1\% & $\pm$\,4.2\% & $\pm$\,1.2\% & $\pm$\,6.4\% & $\pm$\,6.1\%
& $\pm$\,0.7\% \\
\bottomrule
\end{tabular}
\end{table*}

\begin{figure*}[t]
\centering
\includegraphics[width=1.98\columnwidth]{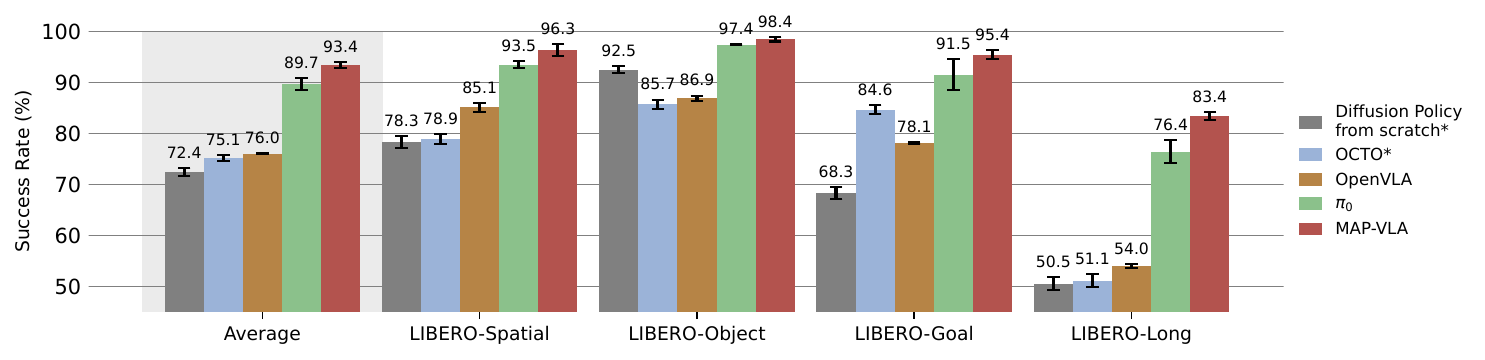} 
\caption{Performance comparison on all LIBERO task suites, ``*'' denotes results reported by OpenVLA~\cite{kim2024openvla}.}
\label{fig:compare1}
\end{figure*}

Using the computed $\alpha_t$, the final memory-augmented action for execution is given by:
\begin{equation}
    \mathbf{A}^{\text{MemAug}}_{t} = \alpha_t \mathbf{A}^{\text{mem}}_{t} + (1 - \alpha_t) \mathbf{A}^{\text{base}}_{t}.
\label{eq:ensemble}
\end{equation}

This prompt ensemble mechanism encourages the policy to favor the action prediction that is closer to the retrieved expert action, effectively leveraging the future-action priors from demonstrations without sacrificing the VLA model’s adaptability to the current scene. By dynamically balancing the task-level generalization of the base prompt with the stage-specificity of the retrieved prompt, the model maintains robustness to retrieval inaccuracies, improves tolerance to stage boundary ambiguity, and preserves adaptability to dynamic test-time scenarios, which results in more stable and accurate behavior during task execution.

In summary, MAP-VLA begins with stage segmentation and tuning stage-specific memory prompts for offline memory construction. The online execution loop then proceeds with (a) observing, (b) retrieving memory, (c) executing dual forward passes, and (d) prompt ensembling via dynamic weighting at each timestep.

\section{EXPERIMENTS}

\subsection{Implementation Details}
\label{sec:imp}
\begin{figure}[t]
\centering
\includegraphics[width=0.99\columnwidth]{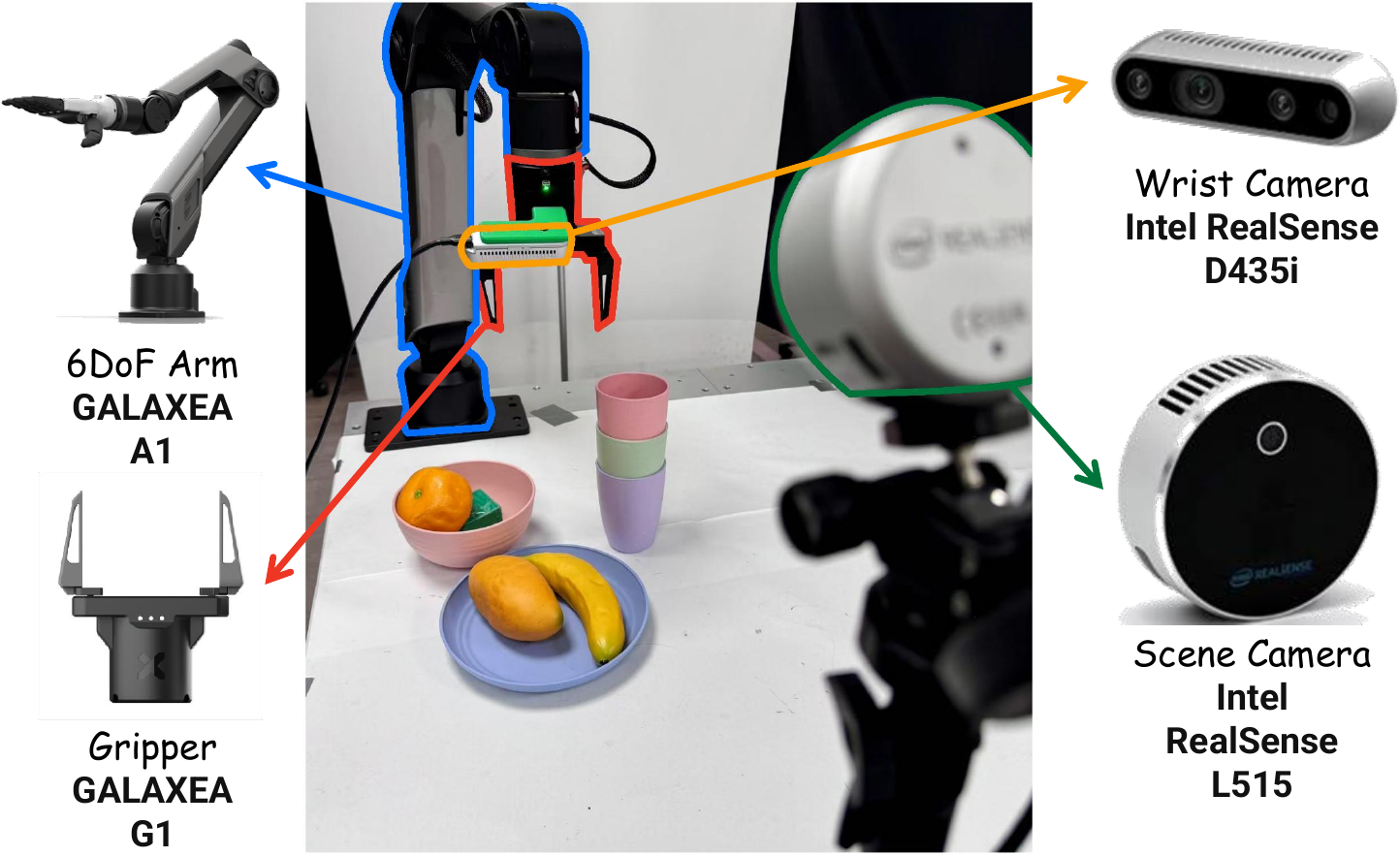} 
\caption{Real-world environment setup.}
\label{fig:real_environment}
\end{figure}

The proposed MAP-VLA is based on the $\pi_0$ model~\cite{black2024pi_0}. We first follow~\cite{black2024pi_0} to fine-tune the $\pi_0$ model on the fine-tuning dataset using LoRA~\cite{hu2022lora} on a server with 6 NVIDIA RTX 6000 Ada GPUs, and then freeze the model weights and apply the same fine-tuning configuration for prompt tuning. For simulation experiments, we use the LIBERO benchmark~\cite{liu2023libero} and adopt the training data and evaluation settings of OpenVLA~\cite{kim2024openvla}, where the success rate is the average over 3 random seeds x 50 rollouts for each task. For real-world experiments, MAP-VLA is deployed on a 6-DoF Galaxea A1 robotic arm shown in Fig.~\ref{fig:real_environment} to perform 20 rollouts for each task. All real-world computations are conducted on a system with an NVIDIA RTX 4090 GPU.

\subsection{Comparison on Long-Horizon Tasks}


\begin{table*}[t]
\centering
\caption{Performance comparison of long-horizon task success rates on real robot evaluations.}
\small
\begin{tabular}{c|cc|cc|cc|cc}
\toprule
\multirow{2}{*}{\textbf{Task}\vspace{-0.6em}} & \multicolumn{4}{c|}{\textbf{Partial Success}} & \multicolumn{4}{|c}{\textbf{Complete Success}} \\
\cmidrule(lr){2-9}
  & $\pi_0$ & MAP-VLA  & Abs. Gain & Rel. Gain & $\pi_0$ & MAP-VLA & Abs. Gain & Rel. Gain \\
\midrule
Task1   &55\%   & \textbf{70\%}& 15\% & 27.3\% &  35\%   & \textbf{50\%} & 15\% & 42.9\% \\
Task2   &55\%   & \textbf{75\%}& 20\% & 36.4\% &  25\%   & \textbf{55\%} & 30\% & 120.0\% \\
Task3   &50\%   & \textbf{60\%}& 10\% & 20.0\% &  10\%   & \textbf{40\%} & 30\% & 300.0\% \\
\midrule
Avg     &53.3\%  & \textbf{68.3\%} & 15.0\% & 28.1\%   &23.3\%  & \textbf{48.3\%} & 25.0\% & 107.1\% \\
\bottomrule
\end{tabular}
\label{tab:subtask-fulltask}
\end{table*}

\begin{figure*}[t]
\centering
\includegraphics[width=1.95\columnwidth]{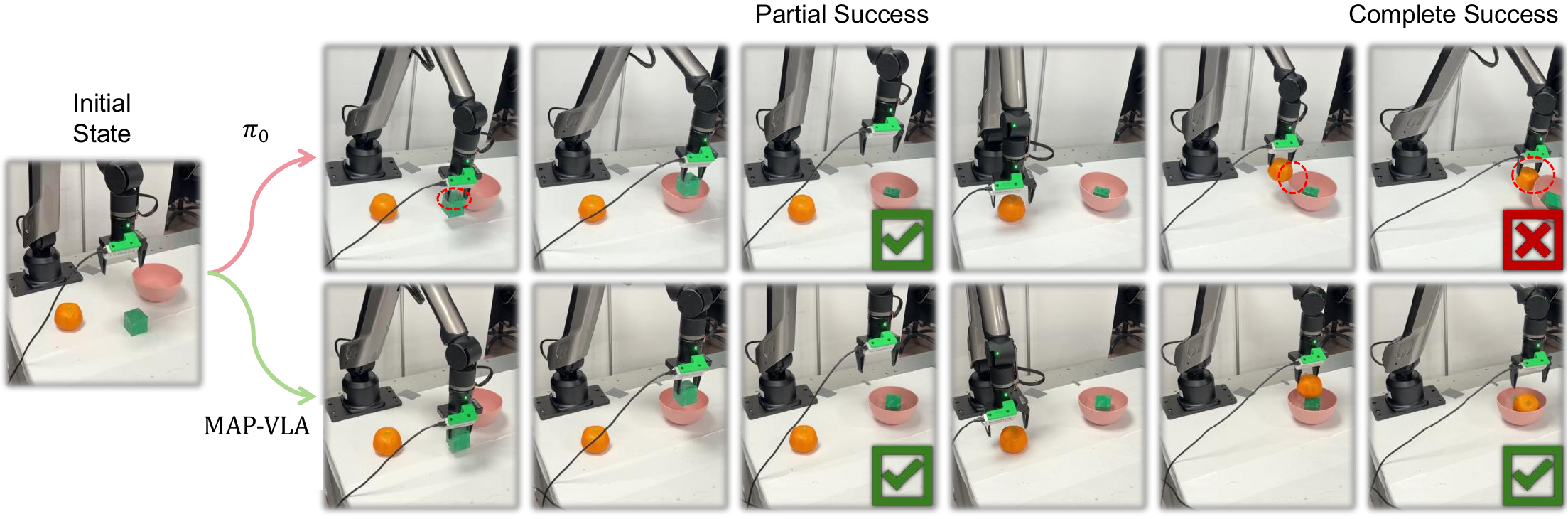} 
\caption{Visualization and comparison of Task2: \textit{place the green cube and orange into the bowl}.}
\label{fig:case_study}
\end{figure*}

\textbf{Comparison on LIBERO simulation}. 
We first compare MAP-VLA to baseline VLA models, OpenVLA~\cite{kim2024openvla} and $\pi_0$~\cite{black2024pi_0}, on the challenging long-horizon manipulation tasks suite LIBERO-Long (also referred to as LIBERO-10) from the LIBERO simulation benchmark. The set includes: (Task1) \textit{pick up the book and place it in the back compartment of the caddy}, (Task2) \textit{put both moka pots on the stove}, (Task3) \textit{put both the alphabet soup and the cream cheese box in the basket}, (Task4) \textit{put both the alphabet soup and the tomato sauce in the basket}, (Task5) \textit{put both the cream cheese box and the butter in the basket}, (Task6) \textit{put the black bowl in the bottom drawer of the cabinet and close it}, (Task7) \textit{put the white mug on the left plate and put the yellow and white mug on the right plate}, (Task8) \textit{put the white mug on the plate and put the chocolate pudding to the right of the plate}, (Task9) \textit{put the yellow and white mug in the microwave and close it}, and (Task10) \textit{turn on the stove and put the moka pot on it}. As shown in Table~\ref{tab:libero-long},  MAP-VLA consistently outperforms all baselines across every individual task in this benchmark. On average, MAP-VLA achieves an 83.4\% success rate, whereas the baseline OpenVLA and $\pi_0$ achieve 54.0\% and 76.4\%, respectively. 
This corresponds to a 7.0\% absolute and 9.2\% relative gains over $\pi_0$, the strongest memoryless baseline. These results underscore the advantage of equipping VLA models with retrievable, stage-specific memory derived from expert demonstrations, particularly in scenarios where long-horizon task decomposition and contextual grounding are critical for success.
We also note that MAP-VLA’s trial outcomes are more consistent, with a lower standard deviation in success rate (0.7\%) across runs than $\pi_0$ (2.3\%). This reduced variability suggests improved robustness and reliability, as a result of encoding additional contextual memory into the prompt and dynamic prompt ensembling as we discuss in Section~\ref{sec:ab}.
We also evaluate MAP-VLA on the full LIBERO benchmark, the results presented in Fig.~\ref{fig:compare1} demonstrate that our method consistently outperforms all baselines across all task suites, highlighting its superior capability beyond long-horizon settings.

\begin{figure*}[t]
\centering
\includegraphics[width=1.99\columnwidth]{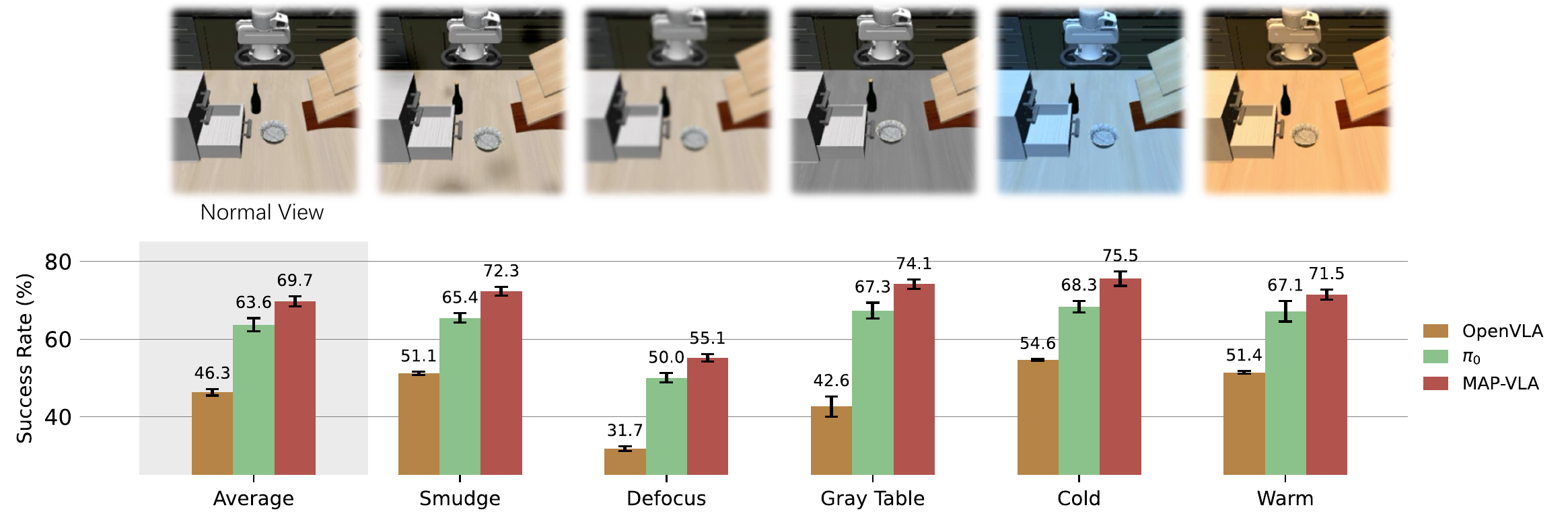} 
\caption{Performance comparison with visual variations on LIBERO-Long.}
\label{fig:compare3}
\end{figure*}

\begin{table*}[]
\centering
\small
\caption{Ablation study on LIBERO-Long.}
\begin{tabular}{l|ccccc}
\toprule
\textbf{Metric} & \textbf{Base VLA} & \textbf{Universal Prompt} & \textbf{Task Prompt} & \textbf{Stage Prompt} & \textbf{MAP-VLA} \\
\midrule
Success Rate (SR) & 76.4\% & 76.9\% & 79.3\% & 81.4\% & \textbf{83.4\%} \\
Standard Deviation (Std) & $\pm$\,2.3\%  & $\pm$\,2.4\%  & $\pm$\,0.8\%  & $\pm$\,2.3\% & $\pm$\,0.7\% \\
\bottomrule
\end{tabular}
\label{tab:ablation}
\end{table*}

\textbf{Comparison on real robot evaluations}. 
To validate the real-world effectiveness of MAP-VLA, we conduct evaluations on a physical robotic platform and compare its performance with the strongest baseline, $\pi_0$, across three long-horizon tasks. Each task comprises two sequential sub-tasks. The evaluated tasks include: (Task1) \textit{place the banana and mango into the plate}, (Task2) \textit{place the green cube and orange into the bowl}, and (Task3) \textit{stack the green cup and pink cup on the purple cup}.
We report both ``Partial Success'' (first sub-task completed) and ``Complete Success'' (both sub-tasks completed). As summarized in Table~\ref{tab:subtask-fulltask}, MAP-VLA again outperforms the baseline policy. Averaged over the three tasks, MAP-VLA’s partial success and complete success rates are 68.3\% and 48.3\%, versus 53.3\% and 23.3\% for the baseline, showing a substantial improvement in overall task completion. 
Notably, the performance gap is more pronounced in complete success in terms of both absolute gains and relative gains, suggesting that memory augmentation plays a particularly critical role in sustaining correct behavior across extended action sequences. Overall, the comparisons confirm that MAP-VLA sets a new state-of-the-art for long-horizon task execution in both simulation and real-robot settings, with significantly higher success rates than prior methods.

\textbf{Case study}. We showcase a specific real-world case to further demonstrate the key improvements of our method. We focus on Task2, where the edges and relative sizes of two objects pose a significant challenge for precise grasping and placement. Such configurations require the policy to distinguish fine-grained spatial cues and execute careful actions. The comparison is visualized in Fig.~\ref{fig:case_study}. Effective long-horizon robot manipulation often demands fine-grained memory to maintain a coherent trajectory across multiple stages. However, the memoryless baseline policy $\pi_0$ exhibits inconsistent and ambiguous object alignment behavior, especially during critical pick-and-place phases (as circled in the figure), often leading to task failure. In contrast, our MAP-VLA framework demonstrates memory-augmented robustness in such settings. By retrieving and incorporating temporally relevant prompts, it enables more stable, context-aware action generation that better adheres to task constraints and results in successful execution under spatial variations.

\subsection{Comparison with Visual Variations}
To assess robustness under real-world visual variations, we evaluate MAP-VLA on LIBERO-Long tasks subjected to various challenging visual conditions. As illustrated in Fig.~\ref{fig:compare3}, we introduce several types of visual perturbations to the robot’s observations during test time: (i) smudge – simulating dirty camera lens; (ii) defocus – simulating an out-of-focus camera; (iii) gray table – a shift in table color from the training demonstrations; (iv) cold – a bluish, low-temperature illumination; and (v) warm – a yellowish, high-temperature lighting condition.
Despite these visual shifts, MAP-VLA consistently retains a significantly higher success rate compared to the baseline $\pi_0$ policy. For instance, under the smudge condition, MAP-VLA maintains 72.3\% success while $\pi_0$ drops to 65.4\%, marking the highest relative gain of 10.6\%. In defocused settings, where fine visual features are blurred, MAP-VLA still outperforms $\pi_0$ (55.1\% versus 50.0\%), aided by memory-based inference from prior demonstrations. MAP-VLA also performs well under table color change, which can be a practical use case, with 74.1\% versus 67.3\%. Overall, MAP-VLA achieves an average relative gain of 9.6\%, slightly above the 9.2\% relative gain without visual variations. Collectively, these findings underscore that memory-augmented prompting remains robust under perceptual variations, suggesting strong generalization capabilities across diverse visual domains.

\subsection{Comparison under Few-Shot Setting}
To further evaluate the adaptability of our method, we compare it with baseline approaches under few-shot learning scenarios. Specifically, the training data contains only 10 or 20 demonstrations per task. 
Table~\ref{tab:fewshot} summarizes the performance of MAP-VLA and $\pi_0$ on the LIBERO-Long benchmark. MAP-VLA achieves average success rates of 55.8\% and 75.9\% for the 10-shot and 20-shot settings, which are consistently higher than those of the baseline $\pi_0$ at 53.6\% and 72.1\%. Moreover, MAP-VLA exhibits lower standard deviation ($\pm$0.9\% and $\pm$0.8\%) compared to $\pi_0$ ($\pm$1.1\% and $\pm$2.0\%), indicating improved robustness even with limited training data.
Overall, MAP-VLA demonstrates strong potential for practical deployment in settings where collecting large-scale demonstrations is infeasible.



\begin{table}[]
\centering
\small
\caption{Comparison under 10-shot and 20-shot on LIBERO-Long.}
\begin{tabular}{l|cc|cc}
\toprule
\multirow{2}{*}{\textbf{Metric}\vspace{-0.6em}} & \multicolumn{2}{c|}{\textbf{10-shot}} & \multicolumn{2}{c}{\textbf{20-shot}} \\
\cmidrule(lr){2-5} 
 & $\pi_0$ & MAP-VLA & $\pi_0$ & MAP-VLA \\
\midrule
SR        & 53.6\%         & \textbf{55.8\%}         & 72.1\%         & \textbf{75.9\%} \\
Std   & $\pm$\,1.1\%   & $\pm$\,0.9\%            & $\pm$\,2.0\%   & $\pm$\,0.8\%    \\
\bottomrule
\end{tabular}
\label{tab:fewshot}
\end{table}

\subsection{Ablation Study}
\label{sec:ab}


We conduct an ablation study on the LIBERO-Long benchmark to quantify the impact of each component in MAP-VLA. Table~\ref{tab:ablation} summarizes the average success rates and standard deviations across five progressively enhanced model variants: (a) Base VLA ($\pi_0$): The original $\pi_0$ model without any prompt tuning or memory augmentation. (b) Universal Prompt: A single soft prompt shared and optimized across all tasks. (c) Task Prompt: A distinct soft prompt per task, applied over the entire trajectory. (d) Stage Prompt: A collection of stage-specific memory prompts, providing fine-grained guidance through retrieval. (e) MAP-VLA: Our complete framework combining stage-specific memory prompts with memory-aware prompt ensembling.

\textbf{Impact of memory-aware prompt ensembling}. The full MAP-VLA (e) variant adds memory-aware prompt ensembling, which dynamically balances the influence of the stage-specific prompt and the base prompt based on their agreement with retrieved demonstration actions. This integration yields the best overall performance, achieving an 83.4\% success rate and the lowest observed standard deviation (0.7\%), reflecting increased consistency and robustness. Crucially, this ensembling strategy enables the model to leverage the complementary strengths of both the task-general base prompt and the stage-specific memory prompts.


\subsection{Retrieval Latency}
\label{sec:com}


In MAP-VLA, memory retrieval is limited to the current and neighboring stages of a task. Assuming task stages are of comparable length, the retrieval cost depends solely on the number of demonstrations for that task and is independent of the total number of tasks and their trajectory lengths. The resulting complexity is \(\mathcal{O}(N)\), where \(N\) denotes the number of training demonstrations for the task. In LIBERO-Long, with an average of 37.8 demonstrations per task, the average retrieval time is only 21.6\,ms.

\section{CONCLUSIONS}
\label{sec:conclusion}
We propose MAP-VLA, a memory-augmented framework that equips a pre-trained Vision-Language-Action (VLA) model with demonstration-derived memory prompts to enhance long-horizon robotic manipulation. By combining prompt tuning with retrieval augmentation, MAP-VLA allows a frozen VLA model to dynamically retrieve and utilize stage-specific memory from expert demonstrations and augment action generation. This flexible and lightweight design improves performance significantly across both simulated benchmark and real-world tasks, achieving higher success rates and increased robustness under visual variations. Our work on memory-augmented VLA contributes positively to the development of more robust and adaptable assistive robots, with potential applications in household support, eldercare, and industrial automation. 
Looking ahead, promising directions for future research include developing generalized yet informative memory prompts that can be reused flexibly across a broad spectrum of tasks, reducing the need for stage-specific memory construction. 







\bibliographystyle{IEEEtran}
\bibliography{ref.bib}

\end{document}